\documentclass[article,onefignum,onetabnum]{siamonline190516}

\usepackage[justification=centering]{caption}
\usepackage{tikz}
\usepackage{pgfplots}
\usepackage{subcaption}
\usepackage{graphicx}
\newcommand{\relu}{\mathrm{relu}}
\usepackage{algorithm}
\usepackage{algorithmic}
\usepackage{comment}

\usepackage{diagbox}

\usepackage{lipsum}
\usepackage{amsfonts}
\usepackage{graphicx}
\usepackage{epstopdf}
\usepackage{algorithmic}
\ifpdf
  \DeclareGraphicsExtensions{.eps,.pdf,.png,.jpg}
\else
  \DeclareGraphicsExtensions{.eps}
\fi

\usepackage{enumitem}
\setlist[enumerate]{leftmargin=.5in}
\setlist[itemize]{leftmargin=.5in}

\newsiamremark{remark}{Remark}
\newsiamremark{hypothesis}{Hypothesis}
\crefname{hypothesis}{Hypothesis}{Hypotheses}
\newsiamthm{claim}{Claim}

\headers{Growing axons: greedy learning of neural networks with application to function approximation}{Daria Fokina \and Ivan Oseledets}

\title{Growing axons: greedy learning of neural networks with application to function approximation\thanks{Supported by RFBR grant 18-31-20069}}

\author{Daria Fokina\thanks{Fraunhofer ITWM, Technical University Kaiserslautern}
  (\email{daria.fokina@itwm.fraunhofer.de})
\and Ivan Oseledets\thanks{Skolkovo Institute of Science and Technology (\email{i.oseledets@skoltech.ru})}}

\usepackage{amsopn}

\ifpdf
\hypersetup{
}
\fi

\begin{document}

\maketitle

\begin{abstract}
 We propose a new method for learning deep neural network models that is based on a greedy learning approach: we add one basis function at a time, and a new basis function is generated as a non-linear activation function applied to a linear combination of the previous basis functions. Such a method (growing deep neural network by one neuron at a time) allows us to compute much more accurate approximants for several model problems in function approximation.
\end{abstract}



\section{Introduction}
Deep neural networks (DNN) have achieved tremendous success in many areas, including image processing, natural language processing, video, and audio synthesis. They have also been used for a long time as a general tool for solving regression tasks, i.e., an approximation of a given function from its samples. Neural networks are known to be a universal approximator for continuous functions \cite{hornik, cybenko}. Recently, several approximation rate results have been established: it has been shown that a certain class of deep neural networks with ReLU \cite{relu} activation functions provide guaranteed convergence rates 
for certain function classes \cite{yarotsky,guhring,liu2019optimal}. Recent paper \cite{opschoor2019deep} provides expressive power results for general piecewise analytic functions with point singularities. These estimates are based on a beautiful example by Yarotsky \cite{yarotsky}, which provides an explicit approximant for a function $f(x) = x^2$ with the convergence that is exponential in the depth of the network.

Surprisingly, there are not too many numerical experiments that construct such approximations using well-developed tools of deep learning. 
In this paper we show experimentally that even for simple functions, such as $f(x) = x^2$ standard algorithms such as stochastic gradient descent (SGD) method fail to converge to high relative accuracy. In order to solve this problem, we propose a new architecture of a deep network, motivated by the explicit construction by Yarotsky, and the new learning algorithm. In terms of neural networks, we incrementally add one neuron in each step in a way it is typically done in the orthogonal matching pursuit in compressed sensing. 
At each substep, we only need to solve a rather simple optimization problem. 
We show experimentally that such an algorithm is able to recover a much better DNN approximation than direct optimization of the L2-norm of the error for the same architecture using standard DNN optimization algorithms.

\section{Approximation of squaring function}
\begin{figure}[!htb]
    \centering
    \includegraphics{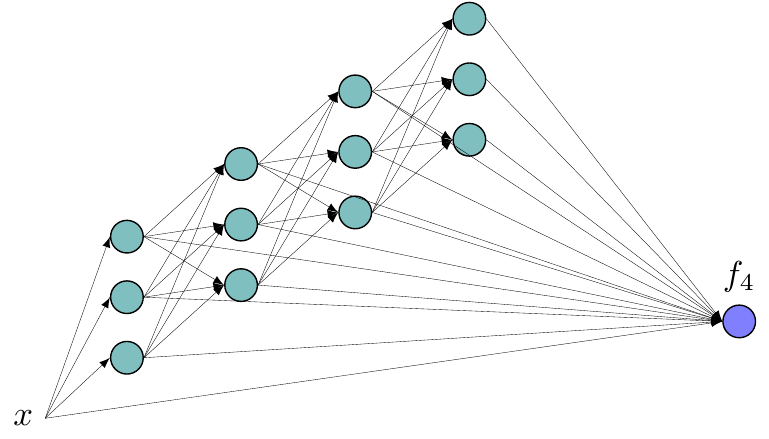}
    \caption{The architecture considered in \cite{yarotsky}}
    \label{axon:fig1}
\end{figure}

\begin{figure}[htb!]
    \centering
    \resizebox{0.9\textwidth}{!}{\input{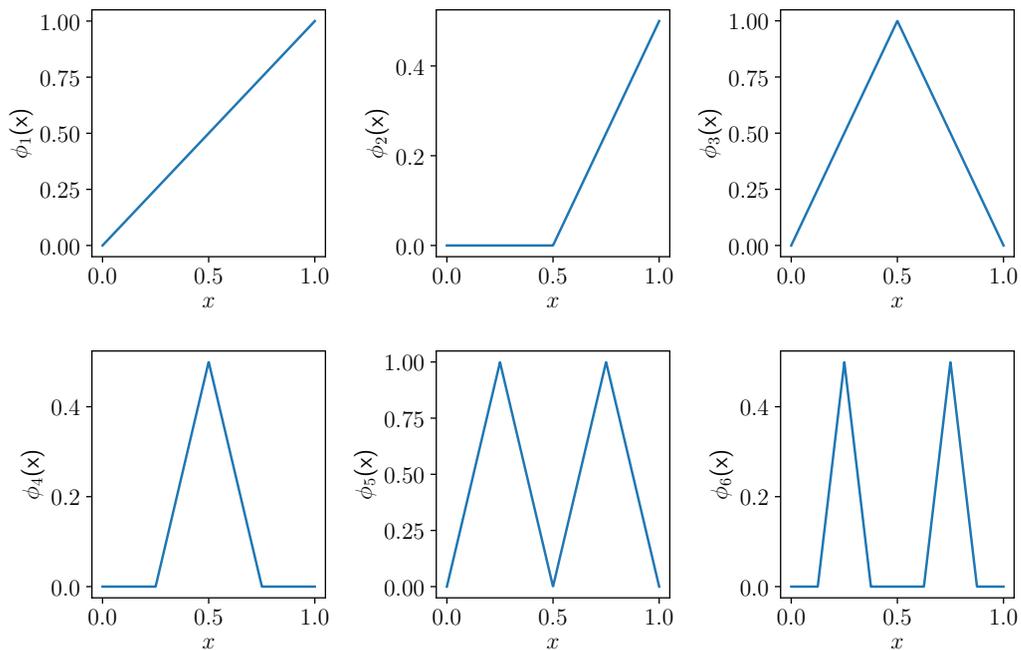}}
    \caption{Basis functions for \cite{yarotsky}}
    \label{axon:basis_functions}
\end{figure}

In order to see the problem, we consider the approximation of a function $f(x) = x^2$. The construction of \cite{yarotsky} has the architecture, shown on Figure~\ref{axon:fig1}. The following Theorem gives the approximation bound.
\begin{theorem}
Suppose
\begin{equation*}
g = \begin{cases} 2x, & 0 \le x < 1/2,\\ 2(1-x), & 1/2 \le x \le 1, \\ 0, & \text{otherwise},\end{cases}
\end{equation*}
\begin{equation*}
g_s = \underbrace{g \circ g \circ..\circ g}_s,
\end{equation*}
then the function
\begin{equation*}
    f_m = x - \sum\limits_{s=1}^m \frac{g_s(x)}{2^{2s}}
\end{equation*}
approximates the function $f(x)=x^2$ with the following error bound:
\begin{equation*}
    |f(x)-f_m(x)| \le 2^{-2m-2}.
\end{equation*}
\end{theorem}
Note, that $g$ can be also rewritten as:
\begin{equation*}
    g(x) = 2\max(x,0)-4\max(x-1/2,0)+2\max(x-1,0),
\end{equation*}
what exactly represents one layer of a fully-connected neural network with $3$ neurons and ReLU activation function.
Note, that this is not a feed-forward network, but a feed-forward network with skip connections: each hidden neuron is connected to the output. Given an input $x$, we can now view the activation at the $i$-th neuron as a \emph{basis function}, which we will denote by $\phi_i(x)$. Some basis functions are shown on Figure~\ref{axon:basis_functions}.

\begin{figure}[htb!]
    \centering
    \resizebox{0.7\textwidth}{!}{\input{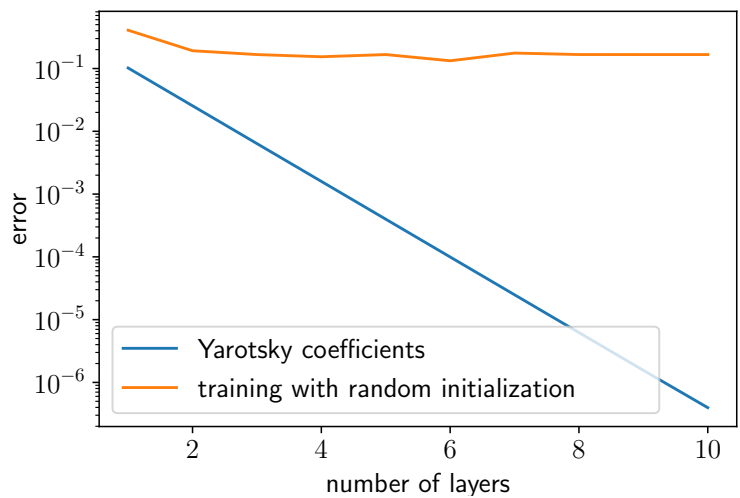}}
    \caption{Comparison of approximation error for weights from \cite{yarotsky} and for the model trained with random initialization}
    \label{axon:errors_ya}
\end{figure}

As we have defined the structure of the network, we can train it from scratch by stochastic gradient descent with random weight initialization. Figure~\ref{axon:errors_ya} shows obtained errors compared to the known explicit construction. 
As one can see, the obtained error value is around $0.1$, while the explicit construction has, as it should, exponential decay.
This example gives a vivid example: the architecture has the expressive power to approximate the function, but it is rather difficult to recover such approximation numerically.

There is one interesting observation. For deep ReLU networks there exists a wide range of approximation error estimates for predefined classes of functions \cite{guhring, perekrestenko, liu2019optimal, Barron1994, Montanelli, yarotsky2018optimal, schmidthieber2019deep, opschoor2019deep}. They are theoretically proven error lower bounds. However, it is not clear how to obtain them using standard deep learning approaches, since the number of parameters is high and the considered optimization problem is non-convex. So, the gradient descent is not likely to reach the global minimum. The convergence of 2-layer neural networks was studied in several works \cite{li2017convergence, zhang2018learning, safran}. One can notice that for different architectures we may obtain different results. For example, it was also shown \cite{safran} that gradient descent with random initialization may converge to a local minimum with probability up to 90\%. On the other hand, there are papers confirming the global convergence \cite{li2017convergence}. For the defined class of loss functions and some prerequisites to the input data it was proven \cite{Zou2019} that SGD with Gaussian initialization of the parameters will also converge to the global minimum. However, our experiments demonstrate the opposite. It might be due to the fact that some prerequisites of the abovementioned results do not hold in our case.

\section{Our algorithm}
\subsection{Learning basis functions}
In order to make a robust learning algorithm, we resort to the basis interpretation of a neural network described above. Let $f(x)$ be the function we are going to approximate. Suppose we have $K$ neurons, and for a given $x$ the activation of the $k$-th neuron is $\phi_k(x)$. In our architecture, every neuron is connected to the output using skip connections, thus the approximation reads
\begin{equation}\label{axon:lstsq}
f(x) \approx f_K(x) = \sum_{k=1}^K c_k \phi_k(x).
\end{equation}
 If we use $L_2$ norm as an error measure between $f(x)$ and $f_K(x)$, the coefficients $c_k$ can be straightforwardly computed from the solution of the linear least squares problem. The main problem is how to learn the functions $\phi_k(x)$. Our proposal is to incrementally add one function at each iteration. Given $K$ basis functions, we would like to introduce a new function $\phi_{K+1}(x)$ that would improve the approximation. 
 \begin{remark}
     The idea of sequentially improving approximation is close to the idea of gradient boosting \cite{friedman2001greedy}: each new \emph{learner} approximates the error. In our case, we approximate not the difference, but the basis functions and coefficients will be recomputed at each step.
 \end{remark}
 
\subsection{Motivation from compressed sensing}
We will build our algorithm upon ideas that are used in compressed sensing. 
Given a matrix $A$ with fewer rows than columns, we want to find a sparse solution $x$ of the underdetermined linear system $Ax = b.$ This can be viewed as the task of selection of optimal basis vectors from the columns of the matrix $A$ such that the target vector $b$ can be approximated by a linear combination of these basis functions, and the number of these basis functions should be as small as possible. The Orthogonal matching pursuit (OMP) method \cite{omp} uses a very simple idea. 
 If we have already selected basis functions $a_{i_1},\ldots, a_{i_K},$
 we can compute the current approximation $x_k$ from the solution of the linear least squares problem by using the QR-decomposition of the matrix
 $$A_K = [a_{i_1}, \ldots, a_{i_K}] = Q_K R_K.$$
 The optimal approximation in the subspace spanned by the columns of $A_K$ is
 $$x_K  = Q_K Q^{\top}_K b,$$
 and the residual is
 $$r_K = (I - Q_K Q^{\top}_K) b.$$
 The next basis function is selected as a column of the matrix $A$ that has the least angle with the residual:
 \begin{equation}\label{axon:ompnextbas}
 i_{K+1} = \arg \max_i \frac{\vert(r_K, a_{i})\vert}{\Vert a_{i} \Vert}.
 \end{equation}
\subsection{Greedy learning of basis functionns}
 This concept can be straightforwardly transferred to our case, but we need to generate possible candidates for a new basis function. We can do that by using a non-linear activation function. For simplicity, let $g(x) = \relu(x) = \max(x, 0)$  be the ReLU activation function. Our proposal is to look for a new basis function in the form
\begin{equation}\label{axon:newbas}
\phi_{K+1}(x) = g(\sum_{k=1}^K (w_K)_k \phi_k(x))
\end{equation}
i.e., we take linear combination of current basis functions, and apply a non-linear activation function. 
In order to formulate the optimization problem for the vector of unknown coefficients $w_K$, 
assume that $x$ takes values only from a sufficiently large training set $x_1, \ldots, x_N$. 
Then, from \eqref{axon:newbas} we get an equation for the new basis vector
\begin{equation}\label{axon:newbasvec}
 v_{K+1} = g( V_K w_K),
\end{equation}
where $V_K$ is an $N \times K$ matrix with elements $(V_K)_{ik} = \phi_k(x_i), i=1, \ldots, N, k=1, \ldots, K$ and $w_K$  is a vector of length $K$ that needs to be determined. 
In order to formulate the optimization problem for $w_K$, we use the same idea as for the OMP method: we minimize the angle between a new basis function and the residual.

\begin{figure}[!htb]
    \centering
    \includegraphics[width=0.8\textwidth]{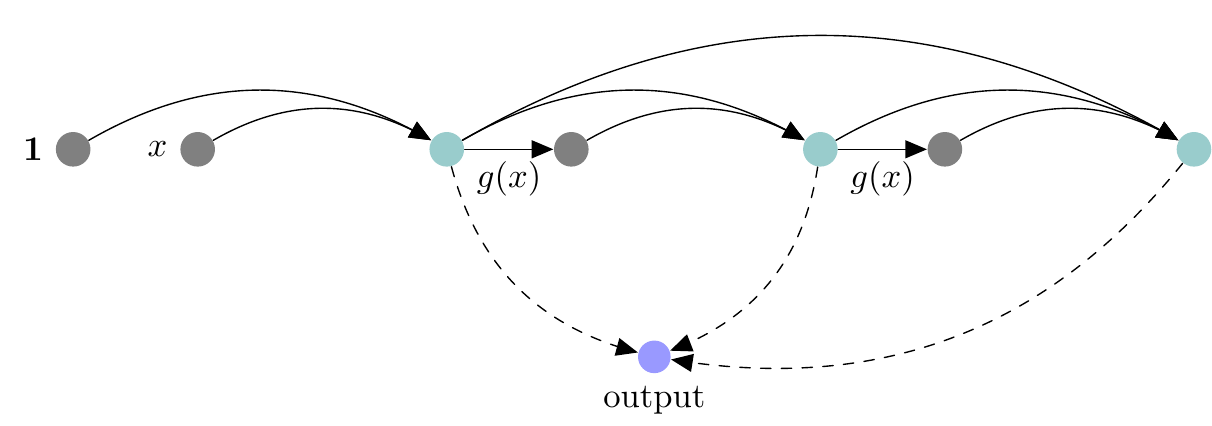}
    \caption{Scheme of Axon network: blue dots are basis functions, skip connections include orthogonalization to the previous functions and normalization operations, for skip connections of the output we can embed them into the coefficients of the linear layer}
    \label{axon:axonarc}
\end{figure}

Let $y$ be the vector with elements $y_i = f(x_i), \quad i = 1, \ldots, N$, i.e., it contains the values of the function we are going to approximate
at the training points. To compute the residual, we first compute the QR-decomposition of the matrix $V_K$:
$$V_k = Q_K R_K,$$
and compute the residual of the best approximation of $y$ by the current basis functions:
$$r_K = (I - Q_K Q^{\top}_K) y.$$
Since $V_K w_K = Q_K R_K w_K = Q_K \widehat{w}_K$ we can write the following optimization problem
for the vector $\widehat{w}_K$:
\begin{equation}\label{axon:optproblem}
    \widehat{w}_K = \arg \max_w \frac{\vert (r_K, g(Q_K w)) \vert}{(w, w)}.
\end{equation}
Note, that if $g$ is the ReLU function, then this optimization problem can be rewritten as
\begin{equation}\label{axon:optproblemsimple}
     \widehat{w}_K = \arg \max_w \frac{\vert (r_K, \vert Q_K w \vert) \vert}{(w, w)},
\end{equation}
since $\relu(Q_K w) = \frac{1}{2} \left( Q_K w + \vert Q_K w \vert \right)$ and
$$
    (Q_K w, r_K) = (w, Q^*_K r_K) = 0.
$$
It would be interesting to find a robust specialized method for solving optimization problems of the form \eqref{axon:optproblemsimple} or \eqref{axon:optproblem}, since it has quite specific structure.
 However, if the number of current basis functions (neurons) is small, then we can resort to efficient global optimization methods. One can also avoid recomputation of the QR decomposition at each step, reducing the complexity. Indeed, if we know $Q_K$ and we have computed $\widehat{w}_K$, we can orthogonalize $\widehat{w}_K$ to the columns of $Q_K$ by using Gram-Schmidt orthogonalization procedure:
$$
   \widetilde{w}_K = \widehat{w}_K - Q_K Q^{\top}_K \widehat{w}_K, \quad \widetilde{w}_K := \frac{\widetilde{w}_K}{\Vert \widetilde{w}_K \Vert}.
$$
This transformation can be parametrized by only one vector of length $K$ and a scaling factor. 
This is the transformation of the basis functions; however, the same transformation will be applied in the process of inference to the values of the neurons. Altogether, in the inference step we need to store two vectors and one scaling factor for each newly added neuron. In fact, in the inference step we can avoid the orthogonalization by absorbing these coefficients into the coefficient $c$, but this may lead to instability.
\begin{remark}
In order to get a more typical neural network architecture, one has to put additional constraints on the vector $w$, i.e., some
of the elements of it should be $0$. While this is not a big problem for functions we consider, we leave this question out of the scope of the current paper: it is a good topic for future research.
\end{remark}
\begin{remark}
Note that the resulting architecture has many connections between neurons. Similar ideas have been previously studied in machine learning in the so-called DenseNet \cite{densenet}, but our architecture is different and we propose an absolutely different learning method. We also note, that to our knowledge, there have been no attempts to apply DenseNet-type architectures to the regression tasks. The orthogonalization step also introduces residual connections: we have a link between all previous neurons and the neuron that stores the value after the activation function has been applied.
\end{remark}
The final ingredient of the method is the initialization since we have to provide some initial basis functions. If the input $x$ has dimension $d$, we propose to use $(d+1)$ functions:
$$\phi_1(x) = x_1, \ldots, \phi_d(x) = x_d, \phi_{d+1}(x) = 1.$$
We will refer to the resulting algorithm and architecture as {\bf Axon} since it has some similarity with the process of growing axons in biological neural networks. The method is summarized in Algorithm~\ref{axon:alg1}.
\begin{algorithm}[!htb]
    \caption{Axon: greedy algorithm for the regression problem}
   \label{axon:alg1}
\begin{algorithmic}
   \REQUIRE  Training set $(x_i, y_i),  \quad x_i \in \mathbb{R}^d, i = 1, \ldots, N$, number of iterations $K$, pointwise nonlinearity $g$.
   \ENSURE Parameters $R, w_k, \alpha_k, \beta_k, k=0, \ldots, K-1, c$ of the neural network.
   \STATE {\bfseries Initialization:} Set $V$ be an $N \times (d+1)$ matrix with elements
   $$V_{i1} = 1, \quad V_{is} = (x_i)_s, \quad s=2, \ldots d, \quad i =1, \ldots, N$$
   \STATE Set $Q, R = \mathrm{QR}(V)$ be the thin QR-decomposition of $V$.
   \FOR{$k=0$ {\bfseries to} $K-1$}
   \STATE Compute residual: $r := y - Q Q^* y$.
   \STATE Solve the minimization problem
       $$w_k = \arg \max_w \frac{\vert (r, g(Q w))\vert}{(Q w, Q w)}.$$
   \STATE Set $w_k := \frac{w_k}{\Vert w_k \Vert}$ and compute the next vector:
          $$\phi := g(Q w_k), \quad \alpha_k = Q^* \phi, \quad q := \phi - Q \alpha_k, \quad \beta_k = \Vert q \Vert, \quad q := \frac{q}{\beta_k}.$$
   \STATE Update basis: $Q := [Q, q]$
   \ENDFOR
   \STATE Compute the coefficients: $c = Q^* y$.
   \RETURN $R$, $\alpha_k, \beta_k, w_k, \quad k=0, \ldots, K-1, c$.
\end{algorithmic}
\end{algorithm}
Once the parameters are computed, we can do the inference with such architecture, as shown in Algorithm~\ref{axon:alg2}. The resulting architecture is shown on Figure~\ref{axon:axonarc}. It can be considered as a neural network with dense and residual connections. It can be very easily implemented in any machine learning framework. After the initialization by Algorithm~\ref{axon:alg1} is computed, we can also fine-tune the whole set of parameters of the Axon network by using SGD-type methods. One can also try to learn such architecture from scratch, but as we will see from the numerical experiments, that correct initialization is crucial.
We also noticed that the complexity of solving optimization subproblem increases for larger number of neurons. We believe that the development of a specialized optimization method that will make use of rather specific structure of this problem can be a very interesting topic for future research.
\begin{algorithm}[htb!]
    \caption{Inference with the Axon architecture}
   \label{axon:alg2}
\begin{algorithmic}
   \REQUIRE  Parameters $R, w_k, \alpha_k, \beta_k, k=0, \ldots, K-1, c$ of the Axon neural network, pointwise nonlinearity $g$, input point $x \in \mathbb{R}^d$
   \ENSURE $y = f_K(x)$
   \STATE Set $v = [x, 1] \in \mathbb{R}^{d+1}$.
   \STATE Set $v := R^{-1} v$.
   \FOR{$k=0$ {\bfseries to} $K-1$}
   \STATE Compute $v_k = \frac{1}{\beta_k}\left(g(w_k^{*} v) - v \alpha_k \right)$.
   \STATE Update $v := [v, v_k]$.
   \ENDFOR
   \RETURN $y = c^{*} v$.
\end{algorithmic}
\end{algorithm}

\section{Numerical experiments}
The implementation of the algorithm is available at \url{https://github.com/dashafok/axon-approximation}. We consider the following model test cases.
\begin{enumerate}
    \item $f(x) = x^2, \quad x \in [0, 1]$,
    \item $f(x) = \sqrt{x}, \quad x \in [0, 1]$,
    \item $f(x) = \exp(-x), \quad x \in [0, 1]$,
    \item $f(x) = \sin(20x), \quad x \in [0, 1]$,
    \item $f(x, y) = \sqrt{x^2 + y^2}, \quad x, y \in [-1, 1]^2$,
    \item Solution of a one-dimensional singularly perturbed reaction-diffusion equation:
    $$-\epsilon^2 u''(x) + u(x) = 1,$$
    $$u(0) = u(1) = 0.$$
\end{enumerate}
\begin{remark}
Our goal is to show that the greedy algorithm is able to learn a good approximation, whereas random initialization with subsequent learning of the parameters of the same architecture by SGD-type method fails. Note, that we have tried different architectures, including classical fully-connected networks, and they show similar performance. 
\end{remark}
Thus, we compare two methods:
\begin{enumerate}
    \item $K$ iterations of Algorithm~\ref{axon:alg1}
    \item Direct optimization of the neural architecture specified by Algorithm~\ref{axon:alg2} with the best loss from $20$ random
        initializations.
\end{enumerate}
We are interested in the approximation accuracy, and also in the basis functions that are learned by our method.

\subsection{Example 1}

The results for $f(x) = x^2$, $x \in [0, 1]$ are shown on Figure~\ref{axon:x2}. First 6 basis functions for the obtained architecture are presented on Figure~\ref{axon:x2_basis}.
\begin{figure}[!htb]
    \centering
    \begin{subfigure}{0.7\textwidth}
    \resizebox{\textwidth}{!}{\input{x2.pgf}}
    \caption{Comparison of relative approximation errors for the axon algorithm and random initializations}
    \label{axon:x2}
    \end{subfigure}
    \begin{subfigure}{\textwidth}
    \resizebox{\textwidth}{!}{\input{basis_x2.pgf}}
    \caption{Basis functions}
    \label{axon:x2_basis}
    \end{subfigure}
    \caption{Relative approximation errors and basis functions for $x^2$}
\end{figure}
\subsection{Example 2}
The results for $f(x) = \sqrt{x}$, $x \in [0, 1]$ are shown on Figure~\ref{axon:sqrt}. First 6 basis functions for the obtained architecture are presented on Figure~\ref{axon:sqrt_basis}.
\begin{figure}[!htb]
    \centering
    \begin{subfigure}{0.7\textwidth}
    \resizebox{\textwidth}{!}{\input{sqrt.pgf}}
    \caption{Comparison of relative approximation errors for the axon algorithm and random initializations}
    \label{axon:sqrt}
    \end{subfigure}
    \begin{subfigure}{\textwidth}
    \resizebox{\textwidth}{!}{\input{basis_sqrt.pgf}}
    \caption{Basis functions}
    \label{axon:sqrt_basis}
    \end{subfigure}
    \caption{Relative approximation errors and basis functions for $\sqrt{x}$}
\end{figure}
\subsection{Example 3}
The results for $f(x) = \exp(-x)$, $x \in [0, 1]$ are shown on Figure~\ref{axon:exp}. First 6 basis functions for the obtained architecture are presented on Figure~\ref{axon:exp_basis}.
\begin{figure}[!htb]
    \centering
    \begin{subfigure}{0.7\textwidth}
    \resizebox{\textwidth}{!}{\input{exp.pgf}}
    \caption{Comparison of relative approximation errors for the axon algorithm and random initializations}
    \label{axon:exp}
    \end{subfigure}
    \begin{subfigure}{\textwidth}
    \resizebox{\textwidth}{!}{\input{basis_exp.pgf}}
    \caption{Basis functions}
    \label{axon:exp_basis}
    \end{subfigure}
    \caption{Relative approximation errors and basis functions for $\exp(-x)$}
\end{figure}
\subsection{Example 4}
\begin{figure}[!htb]
    \centering
    \begin{subfigure}{0.7\textwidth}
    \resizebox{\textwidth}{!}{\input{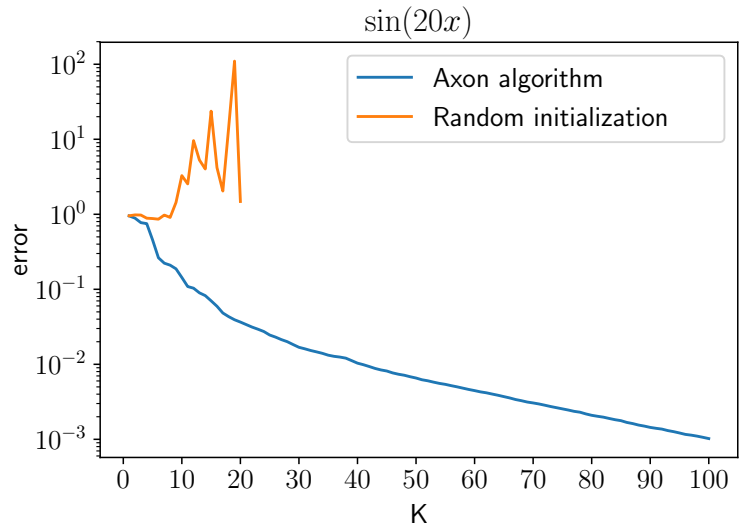}}
    \caption{Comparison of relative approximation errors for the axon algorithm and random initializations}
    \label{axon:sin}
    \end{subfigure}
    \begin{subfigure}{\textwidth}
    \resizebox{\textwidth}{!}{\input{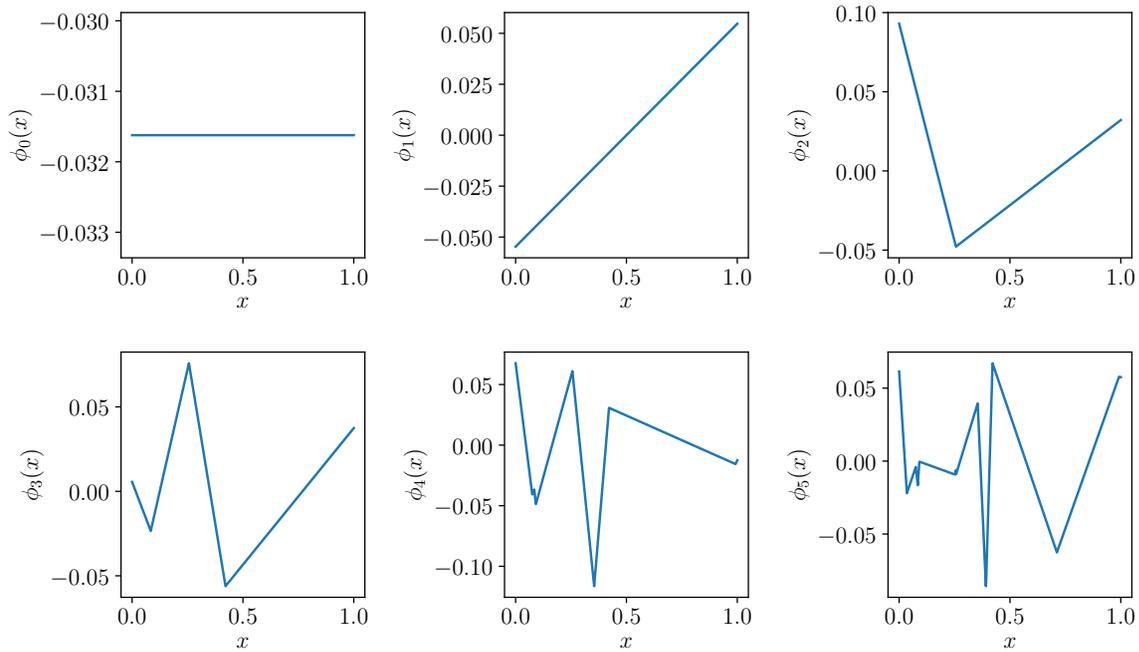}}
    \caption{Basis functions}
    \label{axon:sin_basis}
    \end{subfigure}
    \caption{Relative approximation errors and basis functions for $\sin(20x)$}
\end{figure}

The results for $f(x) = \sin(20x)$, $x \in [0, 1]$ are shown on Figure~\ref{axon:sin}. First 6 basis functions for the obtained architecture are presented on Figure~\ref{axon:sin_basis}.
\subsection{Example 5}
The results for $f(x, y) = \sqrt{x^2 + y^2}$, $x, y \in [-1, 1]$ are shown in Table~\ref{axon:sqrt2d}. First 9 basis functions for the obtained architecture are presented on Figure~\ref{axon:sqrt2d_basis}.
\begin{figure}[!htb]
    \centering
    \begin{subfigure}{0.7\textwidth}
    \resizebox{\textwidth}{!}{\input{sum_squares.pgf}}
    \caption{Comparison of relative approximation errors for the axon algorithm and random initializations}
    \label{axon:sqrt2d}
    \end{subfigure}
    \begin{subfigure}{\textwidth}
    \resizebox{\textwidth}{!}{\input{basis_sum_squares.pgf}}
    \caption{Basis functions}
    \label{axon:sqrt2d_basis}
    \end{subfigure}
    \caption{Relative approximation errors and basis functions for $\sqrt{x^2 + y^2}$}
\end{figure}
    \subsection{Example 6}
As the last example we consider the solution of equation:
    $$-\epsilon^2 u''(x) + u(x) = 1,$$
    $$u(0) = u(1) = 0.$$
It can be written explicitly:
$$u(x) = A\cdot\text{exp}(x/\epsilon)+ B\cdot\text{exp}(-x/\epsilon)+1,$$
where $A = \frac{1-\text{exp}(1/\epsilon)}{\text{exp}(2/\epsilon)-1}$, $B = \frac{\text{exp}(1/\epsilon)-\text{exp}(2/\epsilon)}{\text{exp}(2/\epsilon)-1}$. We consider 2 values of $\epsilon$: $0.1$ and $0.01$. Approximation errors for both $\epsilon$ for different number of neurons are shown on Figure~\ref{axon:diff1} and Figure~\ref{axon:diff2}, correspondingly. First 6 basis functions for the obtained architecture are presented on Figures~\ref{axon:diff1_basis}, \ref{axon:diff2_basis}.
    \begin{figure}[!htb]
    \centering
    \begin{subfigure}{0.7\textwidth}
    \resizebox{\textwidth}{!}{\input{diff_0_1.pgf}}
    \caption{Comparison of relative approximation errors for the axon algorithm and random initializations}
    \label{axon:diff1}
    \end{subfigure}
    \begin{subfigure}{\textwidth}
    \resizebox{\textwidth}{!}{\input{basis_diff_0_1.pgf}}
    \caption{Basis functions}
    \label{axon:diff1_basis}
    \end{subfigure}
    \caption{Relative approximation errors and basis functions for $-\epsilon^2 u''(x) + u(x) = 1, \epsilon=0.1$}
\end{figure}
    \begin{figure}[!htb]
    \centering
    \begin{subfigure}{0.7\textwidth}
    \resizebox{\textwidth}{!}{\input{diff_0_01.pgf}}
    \caption{Comparison of relative approximation errors for the axon algorithm and random initializations}
    \label{axon:diff2}
    \end{subfigure}
    \begin{subfigure}{\textwidth}
    \resizebox{\textwidth}{!}{\input{basis_diff_0_01.pgf}}
    \caption{Basis functions}
    \label{axon:diff2_basis}
    \end{subfigure}
    \caption{Relative approximation errors and basis functions for $-\epsilon^2 u''(x) + u(x) = 1, \epsilon=0.01$}
\end{figure}

Here we must note that the process of training is quite sensitive to the initialization. For the random initialization the error values may grow up to the infinity.
\section{Conclusion and future work}
In this paper we demonstrated that greedy learning can be very efficient for learning neural network architectures compared to standard deep learning approaches: we are able to recover experimentally exponential decay of the error that is predicted by the theory but is quite challenging to verify in practice. We are planning to extend this approach to other tasks, and also use constrained optimization to learn more conventional DNN architectures.

\bibliographystyle{siamplain}
\bibliography{main}

\end{document}